\documentclass{article}

\usepackage[preprint,nonatbib]{neurips_2024}

\usepackage[utf8]{inputenc} 
\usepackage[T1]{fontenc}    
\usepackage{hyperref}       
\usepackage{url}            
\usepackage{booktabs}       
\usepackage{amsfonts}       
\usepackage{nicefrac}       
\usepackage{microtype}      
\usepackage{xcolor}         

\usepackage{amsmath}
\usepackage{graphicx}
\usepackage{multirow}
\usepackage[numbers,square]{natbib}

\title{Watermarking Training Data of Music Generation
Models}

\author{%
Pascal Epple \\ 
 \textit{EPFL} \\
 \And Igor Shilov \\
\textit{Imperial College London} \\
 \And Bozhidar Stevanoski \\ 
\textit{ Imperial College London} \\
 \And Yves-Alexandre de Montjoye\\
 \textit{Imperial College London} \\
}

\begin{document}

\maketitle

\begin{abstract}
Generative Artificial Intelligence (Gen-AI) models are increasingly used to produce content across domains, including text, images, and audio. While these models represent a major technical breakthrough, they gain their generative capabilities from being trained on enormous amounts of human-generated content, which often includes copyrighted material. In this work, we investigate whether audio watermarking techniques can be used to detect an unauthorized usage of content to train a music generation model. We compare outputs generated by a model trained on watermarked data to a model trained on non-watermarked data. We study factors that impact the model's generation behaviour: the watermarking technique, the proportion of watermarked samples in the training set, and the robustness of the watermarking technique against the model's tokenizer. Our results show that audio watermarking techniques, including some that are imperceptible to humans, can lead to noticeable shifts in the model's outputs. We also study the robustness of a state-of-the-art watermarking technique to removal techniques.
\end{abstract}
\section{Introduction}

The rapid advancement of Artificial Intelligence (AI) has revolutionized creative fields, particularly in the generation of multimedia content such as text, images, and audio. In particular, the generation of music using AI models has gained significant attention, with state-of-the-art models capable of generating content that closely resemble human-made compositions \cite{Coscarelli_2023}.

These models gain their generation capabilities from training on vast datasets, which often include copyrighted music created by artists. This raises serious concerns, as these models are prone to memorize and potentially reproduce portions of their training data without proper attribution to the original creators. Such issues have already resulted in numerous lawsuits from content creators against technology companies \cite{Gerken_2024, Knibbs_2024}. As model developers become increasingly reluctant to disclose their training datasets~\cite{bommasani2023foundationmodeltransparencyindex}, new methods to detect the unauthorized use of protected content need to be researched. 

\noindent Watermarking content before its release is a common strategy across multimedia for enabling ownership verification and copyright protection. An audio watermarking technique first embeds a signal into a sample in a way that it is difficult to remove and minimally disruptive to the original content. Then, given an audio, the watermarking technique is able to infer whether it has been watermarked and to extract the embedded signal. Traditional watermarking techniques relied on methods such as Spread-Spectrum Watermarking \cite{spreadspectrum}, or Least Significant Bit Watermarking \cite{lsb}. These methods heavily rely on empirical rules and expert knowledge, and they have not been developed to withstand recently developed audio editing techniques. In recent years, watermarking techniques based on Deep Neural Networks, such as AudioSeal \cite{audioseal} and WavMark \cite{chen2024wavmarkwatermarkingaudiogeneration}, have shown promise. One major advantage to the traditional watermarking techniques is that they can be trained to be robust to a set of pre-defined watermark attacks, while being more imperceptible. 

\noindent In this work, we explore the effectiveness of audio watermarking techniques in enabling the detection of the training data in music generation models. Specifically, we examine whether a content creator can infer unauthorized use of their data by embedding watermarks in their audio before its release. To reproduce this scenario, we fine-tune MusicGen \cite{copet2024}, a state-of-the-art audio generation model, on a dataset containing watermarked samples. We then analyze whether watermarks can be detected in the content generated by the model. We evaluate the effectiveness of watermarking techniques by comparing the output of models trained on watermarked versus non-watermarked datasets. We hypothesize that a model trained on watermarked content will eventually generate watermarked content. 

We begin by evaluating simple and interpretable watermarking techniques made out of tones at different frequencies. Our findings reveal notable shifts in the output distributions of the two models. Our results also highlight the potential of watermarks in the imperceptible range of human hearing frequencies. Furthermore, we show that increasing the percentage of watermarked samples in the training data helps to detect the presence of the watermark in the training data. 

We further consider a more elaborate watermarking technique - AudioSeal. Our results show that content produced by a model trained on samples watermarked with vanilla AudioSeal cannot be differentiated from content produced by a clean model. We hypothesize that this might be due to the watermark not being robust to the model's tokenizer. To address this, we propose iteratively applying AudioSeal—embedding the watermark multiple times in the same audio. This approach improves detectability but comes with trade-offs, as repeated embedding reduces the watermark's imperceptibility.

Overall, our findings demonstrate that watermarking can serve as an effective tool for detecting unauthorized use of content in the training of generative models. Our work takes a first step in developing methods for content creators determine if their data has been used by model developers without their consent. By exploring both simple and more complex watermarking techniques, we provide valuable insights into the trade-offs between robustness, imperceptibility, and effectiveness in protecting creative content.

\section{Prior Work}\label{sec:prior_work}

\subsection{Memorization in Generative Models}

In the context of generative models, memorization can be defined as the model's ability to reproduce or closely mimic specific examples from its training data. Recent studies have shown that generative models, including those for text, image and audio generation, are prone to memorizing training data.

\noindent This phenomenon has been extensively studied in text generation models, such as Large Language Models (LLMs) \cite{carlini2023extractingtrainingdatadiffusion, kandpal2022deduplicatingtrainingdatamitigates, nasr2023scalableextractiontrainingdata}, and image generation models \cite{somepalli2023understandingmitigatingcopyingdiffusion,carlini2023extractingtrainingdatadiffusion, somepalli2022diffusionartdigitalforgery}. These works reveal that text and image-generating models are able to reproduce exact or near-exact copies of training examples.
This raises concerns both from a copyright perspective~\cite{meeus2024copyrighttrapslargelanguage,duarte2024decop}, as well as for privacy and confidentiality~\cite{carlini2019secret,song2017machine}.

\noindent In the audio domain, Barnett et al.~\cite{barnett2024exploringmusicalrootsapplying} explored the concept of "influence attribution" in generative music models. Their work, while not directly addressing exact memorization, provides insights into how generated music can be traced back to influential training samples. Copet et al.~\cite{copet2024} analyzed the memorization abilities of MusicGen, a music generation model they developed. In their work, they found evidence of exact and partial memorization of music samples from the model's training data.

\subsection{Membership Inference}

Model developers are becoming increasingly reluctant to reveal details of their training sources \cite{bommasani2023foundationmodeltransparencyindex}. Recent research has introduced methods to perturb the training data of Machine Learning models, enabling the detection of whether a model has been trained on it.

Membership Inference Attacks (MIAs) have been proposed to study the level of memorization exhibited by an ML model~\cite{shokri2017membership, ye2022enhanced, carlini2022membership}. In an MIA, an adversary aims to determine whether a given data sample was used to train a model or not.

In many real-world scenarios, however, standard MIA are challenging to apply - due to prohibitive costs of training shadow models~\cite{carlini2022membership,ye2022enhanced}, data distribution shift between members and non-members~\cite{meeus2024sokmembershipinferenceattacks, das2024blindbaselinesbeatmembership} or simply a low natural memorization of a targe model~\cite{meeus2024copyrighttrapslargelanguage}. Methods have been proposed to modify the training data - either by adding new training samples or amending the existing ones - to enhance the MIA performance in detecting the training samples.

\noindent In the image domain, Sablayrolles et al.~\cite{sablayrolles2020radioactivedatatracingtraining} introduced a technique to infer whether a subset of data samples has been seen by an image classifier during training. They propose a method called "radioactive data", which involves adding imperceptible, unique perturbations to part of the dataset, allowing them to later determine whether the classifier has been trained on it. Similarly, researchers have been focusing on text-to-image diffusion models. Wang et al.~\cite{wang2024diagnosisdetectingunauthorizeddata} propose an approach, referred to as DIAGNOSIS, which coats images by adding an imperceptible perturbation to a subset of images in the training data. The method enables to infer whether a diffusion sample has memorized these perturbations, providing a robust mechanism to protect the content.

\noindent In the text domain, text watermarks, also known as canaries~\cite{carlini2019secret,mireshghallah-etal-2022-empirical} or copyright traps~\cite{meeus2024copyrighttrapslargelanguage} have been proposed to exploit the memorization abilities of language models. These traps consist of synthetically generated text injected into the original content of a document. When the trap sequences are repeated a large number of times throughout a document, they enable inferring whether the document has been seen by an LLM during training. However, copyright traps remain vulnerable to sequence-level de-duplication, one of the most common pre-processing steps used to ensure high-quality training data.

\section{Method}\label{ch:method}

\subsection{Training Data Watermarking}

Given a training dataset $D$ of music samples and a watermarking technique $W$, we watermark a proportion $p$ of the training samples with $W$, and leave the remaining samples untouched. We refer to a dataset containing watermarked samples as a \textit{watermarked dataset}. Similarly, a \textit{clean dataset} refers to a dataset that does not contain any watermarked samples.

\subsection{Training Two Models}
Our objective is to infer whether it is possible to differentiate between a model trained on watermarked data and a model trained on clean data. To this end, we train one instance of a music generation model $M$ on the clean dataset, and another one the watermarked dataset. The two models are trained with the same training procedure and hyperparameters. We refer to a model trained on clean data as a \textit{clean model}, and to a model trained on watermarked data as a \textit{watermarked model}.  

\subsection{Generating Continuations of Watermarked Audios}
We create a dataset $O$ of 10-seconds long continuations of audio prompts. For each audio prompt, we generate a continuation by the clean and by the watermarked model. The audio prompts consist in the first $d$ seconds of samples from the model's validation dataset. This allows us to attribute any differences in the outputs to the injected watermark, which is the only difference between the clean and the watermarked models, rather than for example to the memorization of training samples. The audio prompts are watermarked with the same technique $W$ as used to watermark the training data. With $N$ prompts, we create a dataset of $2N$ outputs and label the output according to the model that generated it. As text prompt, we use the original caption provided in the dataset.

\subsection{Model Attribution of the Outputs}

We use a binary classifier $C$ to predict whether an output was produced by the watermarked or by the clean model. The classifier is pre-trained and corresponds to the detector of the watermarking technique. If the generated outputs can accurately be attributed to the model that produced it, there is solid evidence that the watermark has an impact on the generation behavior of the model. 

\noindent We evaluate the classifier by computing the Area Under the ROC Curve (AUC). The AUC is a metric that quantifies the overall ability of the classifier to distinguish between classes. It is a threshold independent metric that ranges from 0 to 1. An AUC of 0.5 corresponds to random guessing.

\subsection{Watermark Imperceptibility}
Following previous work \cite{audioseal, chen2024wavmarkwatermarkingaudiogeneration}, we use the Scale-Invariant Signal to Noise Ratio (SI-SNR) and Perceptual Evaluation of Speech Quality (PESQ) as evaluation for the quality of watemraked audios. Given an audio sample $s$ and its watermarked version $s_w$, the formula to compute SI-SNR is as follows:
\begin{equation}
\text{SI-SNR}(s, s_w) = 10 \log_{10} \left( \frac{\lvert \alpha s \rvert_2^2}{\lvert \alpha s - s_w \rvert_2^2} \right), \textit{where } \alpha = \frac{\langle s, s_w \rangle}{\lvert s \rvert_2^2}
\end{equation}
When the SI-SNR is high, the watermark is less perceptible. 

\noindent The PESQ is designed to assess the quality of audio as perceived by human listeners. PESQ scores range from -0.5 to 4.5, with scores above 4 indicating good auditory quality. The PESQ scores are then rescaled with a calibration technique to Mean Opinion Scores (MOS). The MOS cover a scale from 1 to 5, where 1 means bad audio quality, and 5 means excellent audio quality. We report the re-calibrated PESQ scores, and refer to them as PESQ. When the PESQ score is high, the watermark is less perceptible.  

\noindent We use these two objective metrics to compare the imperceptibility of AudioSeal on the two datasets. While we believe that these metrics are already solid first indicators, they are not sufficient to fully evaluate imperceptibility. Indeed, there is no pre-determined threshold, nor for SI-SNR or for PESQ, that determines whether the watermark is perceptible or not. This is one of the many reasons why studies on imperceptibility often involve human evaluations. These are out of the scope of this work.
 
\subsection{Impact of Watermark on Model Performance}
Depending on the watermarking technique used, it may significantly alter the training data of the music generation model, and thus significantly corrupt the performance of the model. In line with previous work \cite{copet2024}, we compute two metrics: the Fréchet Audio Distance and the Kullback-Leibler Divergence over a music classification model.

\subsubsection{KL-Divergence over PaSST Classifier}
The evaluation metric is based on the Kullback-Leibler divergence (KLD), a statistical measure that compares two data distributions. Specifically, given two probability distributions $P$ and $Q$, the KLD is computed as follows:

\begin{equation*}
    D_{KL}(P \parallel Q) = \sum_{x \in \mathcal{X}} P(x) \log\left(\frac{P(x)}{Q(x)}\right)
\end{equation*}

\noindent The KLD is asymmetric, meaning \( D_{KL}(P \parallel Q) \neq D_{KL}(Q \parallel P) \). We always report the minimum KLD value.

\noindent The evaluation metric is based on a state-of-the-art audio classifier: the Patchout Fast Spectrogram Transformer (PaSST). This model has been trained to accurately tag audio clips into 527 possible classes that best represent the audios. Examples of such classes are "drum", "singing" or "electric piano". 

\noindent The evaluated model generates a music sample from a text caption associated to a reference music sample. We then compute the probabilities of the classification labels for both music samples, and compute the KLD over these probability distributions. A low KLD suggests that the generated audio shares similar concepts (melody, rhythm, timbre, etc.) with the reference audio.  

\subsubsection{Fréchet Audio Distance}
The Fréchet Audio Distance (FAD) \cite{frechetaudiodistancemetric} is a metric designed to evaluate the quality of audio. FAD does not evaluate individual audios, but compares the distributions of features extracted from a large set of clean and real audio samples (e.g. the training set) and from a set of generated audios. One advantage of FAD is that it does not rely on a ground truth of the evaluated audios, and is thus a reference-free metric. FAD also correlates with human perception of audio quality.

A pre-trained audio feature extractor is used to extract features that capture important characteristics of the audio. FAD uses the VVGish model \cite{vggish}. The embeddings of the evaluation set and of the training set are then represented as multivariate Gaussians, $\mathcal{N}_r(\mu_r, \Sigma_r)$ and $\mathcal{N}_g(\mu_g, \Sigma_g)$, respectively. The Fréchet distance between these two Gaussians is computed as follows:
\begin{equation}
\textbf{F}(\mathcal{N}_r, \mathcal{N}_g) = \lvert \mu_r - \mu_g \rvert + \text{tr}(\Sigma_r + \Sigma_g - 2\sqrt{\Sigma_r \Sigma_g})
\end{equation}
where $tr$ denotes the trace of a matrix. The FAD has been designed to correlate with human perception of sounds. A low FAD scores suggests that an audio is more plausible.

\section{Experimental Details}
\label{ch:experiments}

\subsection{Dataset}\label{sub:musiccaps}
The MusicCaps \cite{agostinelli2023musiclmgeneratingmusictext} dataset contains music examples that are labeled with an aspect list, and a textual description of the audio written by musicians. An aspect list is for example "low quality, noisy, mono, acoustic rhythm guitar, flat male vocal, passionate, country", while the description is more detailed, e.g. "The low quality recording features a flat male vocal singing over acoustic rhythm guitar chords, after which that same vocal is talking. The song sounds passionate, while the recording, overall, is noisy and in mono." The music samples are 10 seconds long, and sampled at 48kHz.

\noindent The samples in MusicCaps stem from segments of YouTube videos that contain music. A publicly available tabular dataset includes the YouTube video IDs, along with the start and end timestamps of the music samples within each video. We manage to download 5402 out of the total 5521 segments listed in the dataset. The links to some videos have expired, and others went private, and can thus no longer be downloaded. For instance, the YouTube video with ID "0J\_2K1Gvruk" is no longer accessible because it went private.

\subsection{Considered Watermarks}\label{sec:injected_watermarks}

We study two types of watermarks: tone-based watermarks and AudioSeal-based watermarks. AudioSeal \cite{audioseal} is a state-of-the-art watermarking technique. It generates an additive watermark which is imperceptible to humans. Tone-based watermarks serve the purpose of an initial proof-of-concept, and on the memorization capabilities of the model. They follow simple and interpretable rules.

\subsubsection{Tone-based Watermarks}

The tone-based watermarks are made out of pure cosine tones at a frequency $f$. We here introduce different tone-based watermarks and some notations:

\begin{itemize}
    \item \textbf{Tone $f$ :} We add a tone at a frequency of $f$ Hz to the audio, covering the entire range of the audio.
    \item \textbf{Switch $d$}: We add a tone at frequency $f$ for a duration $d$, and then switch the frequency to a frequency $f'$. The watermark covers the entire audio.
    \item \textbf{Alternate $d$}: The watermark consists of alternating tones at frequency $f$ and $f'$. Each tone is of duration $d$, and the watermark covers the entire audio.
    \item \textbf{Stop $d$}: In the first $d$ seconds of the audio, we add a tone at frequency $f$ for a duration $d$. The rest of the audio is not affected by the watermark and remains the same.
\end{itemize}

\begin{figure}
    \centering
    \includegraphics[width=1\linewidth]{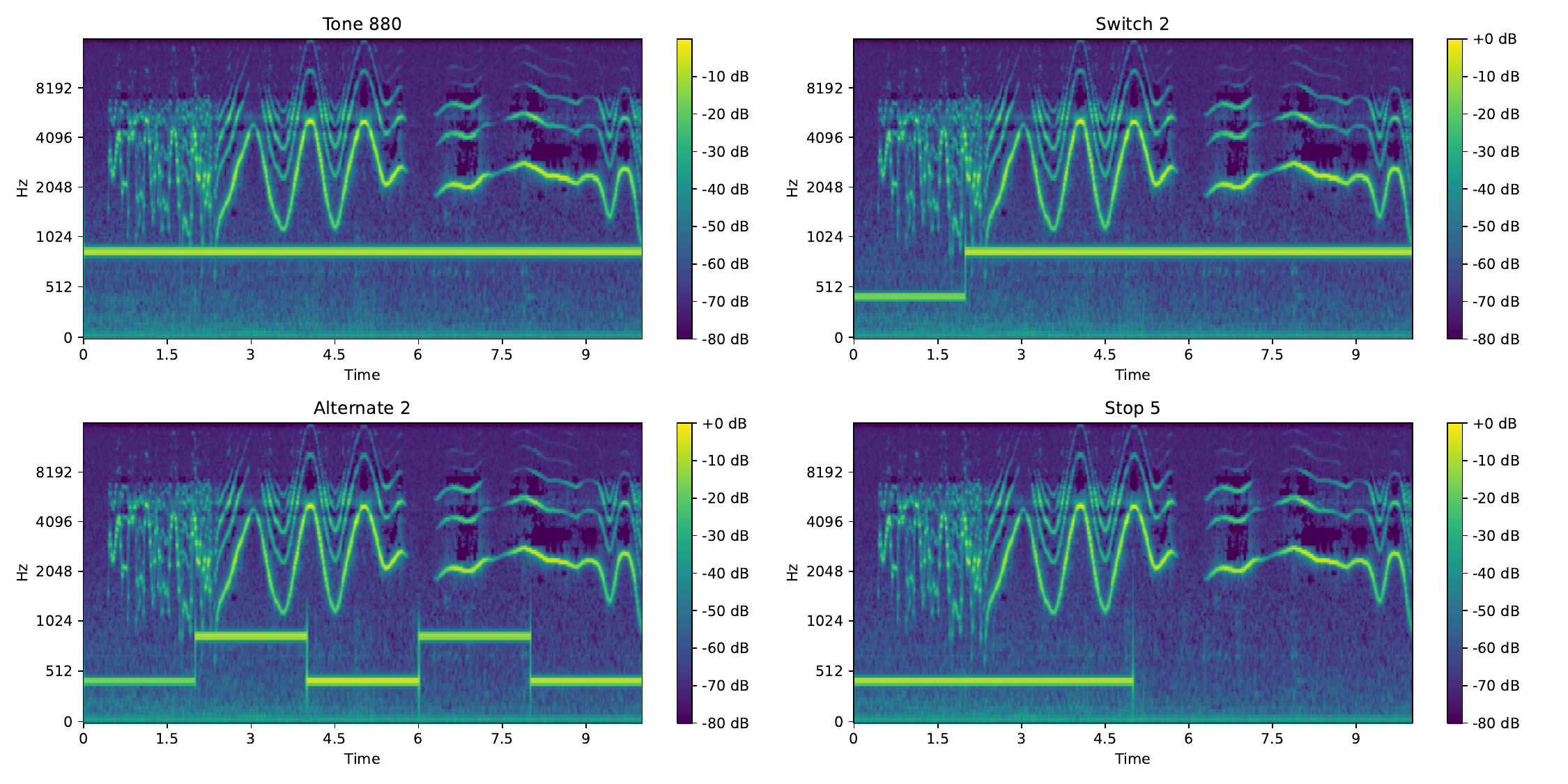}
    \caption{\textbf{Mel-Spectrograms of Watermarked Audio: } The spectrograms of an audio sample watermarked with the following techniques: Tone 880, Switch 2, Alternate 2 and Stop 5. The frequencies $f$ and $f'$ are set to 440 Hz and 880 Hz, respectively.}
    \label{fig:specs_wm_techniques}
\end{figure}

Figure \ref{fig:specs_wm_techniques} illustrates the spectrogram of an audio file watermarked with different tone-based watermarks. In the objective to make the injected watermarks more subtle and less perceptible, we re-scale the tones with the Root Mean Square (RMS) amplitude of the original audio signal before adding them to the signal. 

\noindent Given an audio of duration $d$ watermarked with \textit{Switch d} or \textit{Alternate d}, it is impossible to guess the continuation of the audio without knowledge of the watermark. This is why we refer to these watermarking techniques as \textit{Secrets}.

The human ear can detect sounds in a frequency range from about 20Hz to 20kHz \cite{purves2001neuroscience}. We aim to take advantage of this by injecting tones in the imperceptible range of frequencies in the audios. Remember that MusicGen is trained on audios sampled at 32 kHz, and thus have a maximum frequency of 16 kHz. This leaves a narrow, yet existing band of frequencies between 0 Hz and 20 Hz of imperceptible frequencies.

\subsubsection{AudioSeal-based Watermarks}
We consider the following watermarks based on AudioSeal:
\begin{itemize}
    \item \textbf{AudioSeal:} We watermark the audio samples using AudioSeal.  
    \item \textbf{Multi $k$:} We watermark the audio samples $k$ times using AudioSeal.
\end{itemize}

These watermarks are unique to each audio sample, as AudioSeal generates a different watermark for every audio file to maximize imperceptibility.

To enhance robustness to MusicGen's tokenizer, we use \textit{multiple watermarking} which we here define. Given an audio sample $s$, we watermark it $n$ times with AudioSeal $A$ in the following way:
\begin{equation}
    s_a = \underbrace{A \circ \cdots \circ A}_{\text{$n$ times}}(s)
\end{equation}
To detect the presence of the watermark in $s_a$, we use AudioSeal's default detector.

\subsection{Model, Data Processing and Fine-Tuning Details}

A thorough study on the effect of different watermarks requires to train many different music generation models. To keep compute tractable, we decide to fine-tune a pre-trained model on a smaller dataset. Throughout all of our experiments, we use MusicGen-small as music generation model, and refer to it as \textit{MusicGen}. We refer to the audio tokenizer, or auto-encoder, used by the the model as Encodec 32.

The architecture of MusicGen consists of an auto-encoder that encodes audio into discrete tokens, and a single-stage transformer that predicts the next tokens. In the released models, the auto-encoder is EnCodec \cite{défossez2022highfidelityneuralaudio}, but the model developers have successfully tested their model with Descript-Audio-Codec \cite{kumar2023highfidelityaudiocompressionimproved} as well. Finally, the auto-encoder decodes the generated stream of tokens back into audio.

\noindent We fine-tune MusicGen on samples from MusicCaps, the dataset introduced in \ref{sub:musiccaps}. Since MusicCaps is part of the evaluation data of MusicGen, we are confident that it has not yet been seen by the model.

\noindent Here, we detail the pre-processing steps applied to the dataset to prepare it for the fine-tuning process. First, we resample the 10 second long audio samples from 48 kHz to 32 kHz. As MusicGen has been originally trained on 30 second samples, we extend the audios by repeating them 3 times. This has led to significantly better fine-tuning performances than zero-padding the audios. The size of MusicCaps is roughly 1000 times smaller than the dataset used to pre-train MusicGen. Regarding the metadata information of the audios, we only consider the raw textual description of the music samples already provided in MusicCaps. To evaluate the performance of the fine-tuned models, we build a test set consisting of 50 samples. The remaining samples are split into a training set and a validation set using a traditional 80-20 split. The training, validation and evaluation splits are the same for every fine-tuned model.

\noindent Each MusicGen model in this work has been fine-tuned in the exact same manner as we here describe, with the only difference being the training dataset, which can contain watermarked samples. We fine-tune the 300M parameter model on cross-entropy for 5 epochs, with 250 updates per epoch, and a batch size of 16. This process takes around 25 minutes on 2 NVIDIA A-100 GPUs. We use the default hyper-parameters (learning rate, optimizer, etc.) provided by the model developers, as those led to good training performance. We use the delay pattern as the decoding strategy. This is the same strategy as used in the released models. During model fine-tuning, the audio tokenizer (Encodec 32) and the text encoder (T5) are both frozen.

\subsection{Generating Outputs from the Models}

\noindent The audios are generated from 200 randomly chosen samples from the validation set of MusicCaps. Each one gets watermarked with the same technique as used to train the watermarked model, and shortened to a duration of $d$ seconds. We refer to this shortened, watermarked audio as the \textit{audio prompt}. For each audio prompt, we generate a continuation from the watermarked model and from the clean model. We use the raw textual description provided in the dataset as text prompt. We use the default generation parameters of the model: top-250 sampling, with a temperature of 1. The final dataset is made out of 400 samples, equally balanced between music generated by the watermarked and by clean model. We evaluate our method on three different datasets constructed by modifying the sampling seed of both models.

\subsection{Binary Classifiers}
In attempt to separate between audios generated by the clean and the watermarked model, we use the detectors of the watermarking techniques. This guarantees interpretability of the results. For AudioSeal-based watermarks, we use its detector to predict scores that indicate whether the sample is watermarked. To accurately detect the watermarks based on tones, we implement a rule-based classifier.

\subsubsection{Rule-Based Classifier}
The objective of the rule-based classifier is to detect the presence of a tone at a frequency $f$ in an audio file. It is based on the mel-spectrograms of the audio, computed with 128 Mel bands, a hop length of 512 and an FFT window length of 2048. Then, we identify the Mel band with the center frequency closest to the target frequency $f$. The intuition behind the rule-based classifier is that, in the presence of a tone at this specific frequency in the audio, the corresponding Mel band contains higher values compared to when the tone is absent. The prediction of the classifier is computed by summing the values in the Mel band. When using watermarks that involve two different frequencies $f$ and $f'$ (e.g. \textit{Secrets}), we need to make sure that they fall in different Mel bands.

\section{Results}\label{ch:results}

\subsection{Robustness of AudioSeal to MusicGen's Tokenizer}

\begin{figure}
    \centering
    \includegraphics[width=0.8\linewidth]{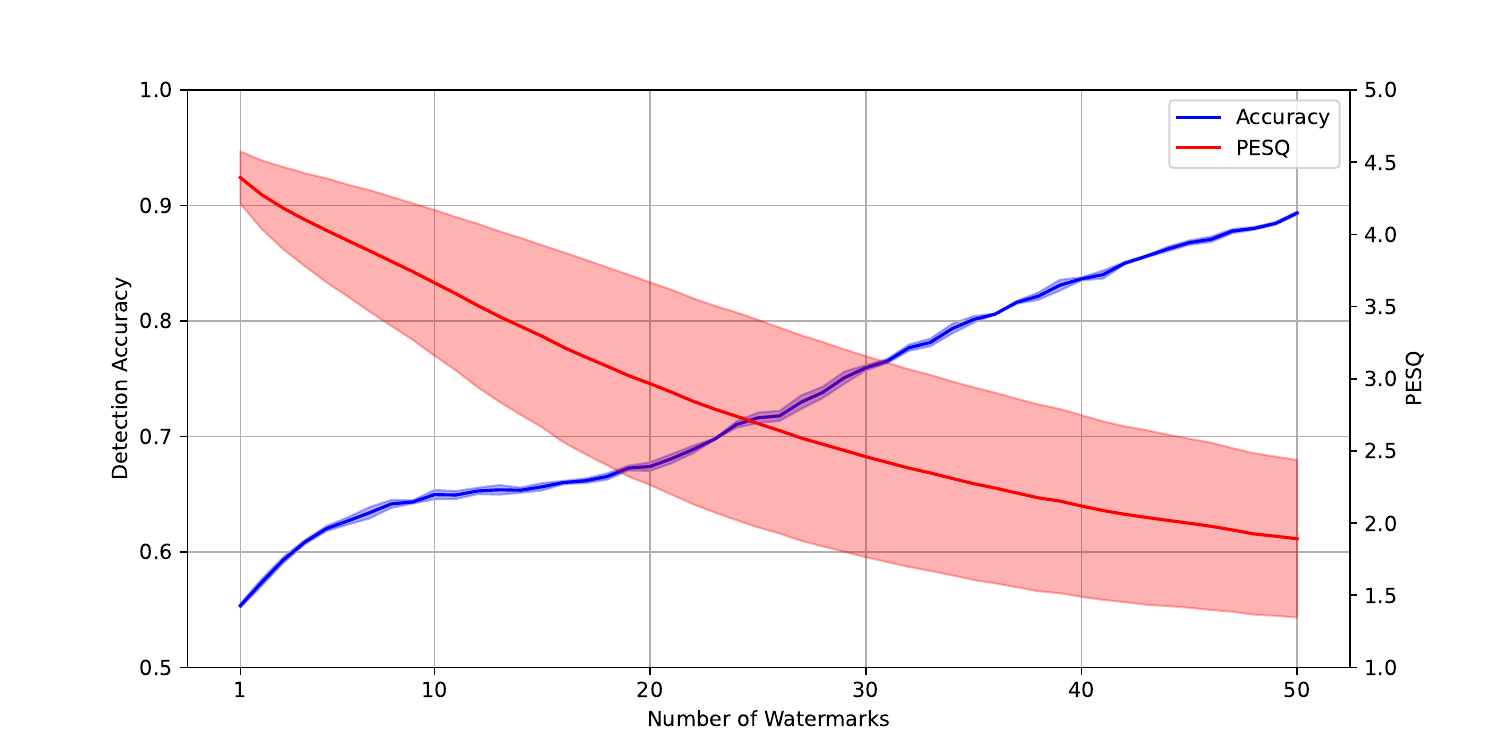}
    \caption{\textbf{Detection Accuracy with Multiple Watermarking:} We compute the detection accuracy on compressed audio samples from MusicCaps when applying AudioSeal multiple times, after Encodec32 compression.} 
    \label{fig:mult_wm_detection}
\end{figure}

We first investigate the robustness of AudioSeal to Encodec32, the audio tokenizer of MusicGen. Figure \ref{fig:mult_wm_detection} shows that AudioSeal is not robust to Encodec32 as the detector only achieves a detection accuracy of 0.5742. Hence, we explore the robustness when watermarking the sample multiple times with AudioSeal. We watermark the samples up to 50 times, and still use AudioSeal's detector to compute a detection score on the compressed samples. In Figure \ref{fig:mult_wm_detection}, we notice a monotonic increase in the detection accuracy with respect to the number of watermarks applied. The detection accuracy rapidly increases between 1 and 10 number of watermarks. When watermarking the audio samples 50 times, AudioSeal's detector achieves a detection accuracy of 0.8935. This is a significant increase to the detection accuracy of vanilla AudioSeal. It is worth to note the detector has never been exposed to such watermarked data, yet accurately detects the presence of the watermark. This suggests that the encoded watermark still follows the principals of the watermarking technique when it is applied once: the encoded watermark does not result in random noise added to the audios. We have successfully built an AI-based watermarking technique that is robust to Encodec32 compression. However, this comes at the price of imperceptibility. As shown in Figure \ref{fig:mult_wm_detection}, the PESQ of the watermarked audios indicates a severe degradation in imperceptibility when increasing the number of watermarks applied. The watermark remain of "fair" quality up to 10 consecutive applications of AudioSeal. For 50 applications, the audios are rated as "poor" and "bad". Our results highlight the clear trade-off between imperceptibility and robustness to the audio tokenizer. 

\begin{table}[ht]
    \centering
    \vspace{0.2cm}
    \resizebox{1\linewidth}{!}{
        \begin{tabular}{lcccccc}
            \toprule
            \multirow{2}{*}{\textbf{Watermark}} & \multirow{2}{*}{\textbf{AUC} ($\uparrow$)} & \multicolumn{2}{c}{\textbf{Imperceptibility}} & \multicolumn{2}{c}{\textbf{Model Utility}} \\
            \cmidrule(lr){3-4} \cmidrule(lr){5-6}
            & & \textbf{PESQ} ($\uparrow$) & \textbf{SI-SNR} ($\uparrow$) & \textbf{FAD} ($\downarrow$) & \textbf{KLD} ($\downarrow$) \\
            \midrule
            None & 0.5 & N/A & N/A & 5.8817 $\pm$ 0.5616 & 1.6709 $\pm$ 0.1173 \\
            \midrule
            \multicolumn{6}{l}{\textbf{Tone-based}} \\
            Tone 10 & 0.7112 $\pm$ 0.0359 & 4.59 $\pm$ 0.18 & 3.00 $\pm$ 0.03 & 5.9108 $\pm$ 0.2911 & 1.6632 $\pm$ 0.1183 \\
            Tone 440 & 0.7787 $\pm$ 0.00091 & 2.12 $\pm$ 0.48 & 3.00 $\pm$ 0.03 & 6.0456 $\pm$ 0.5855 & 1.5857 $\pm$ 0.0805 \\
            Stop 5 & 0.6426 $\pm$ 0.0097 & 3.60 $\pm$ 0.43 & 11.34 $\pm$ 2.68 & 5.9750 $\pm$ 0.2646 & 1.6631 $\pm$ 0.0222 \\
            Secret 5 & 0.5724 $\pm$ 0.0205 & 1.86 $\pm$ 0.46 & 3.00 $\pm$ 0.03 & 5.7438 $\pm$ 0.1932 & 1.5762 $\pm$ 0.0690 \\
            \midrule
            \multicolumn{6}{l}{\textbf{AudioSeal-based}} \\
            AudioSeal & 0.4914 $\pm$ 0.0157 & 4.39 $\pm$ 0.18 & 25.15 $\pm$ 4.23 & 5.7603 $\pm$ 0.4686 & 1.6556 $\pm$ 0.2178 \\
            Multi 10 & 0.5138 $\pm$ 0.0034 & 3.51 $\pm$ 0.51 & 14.17 $\pm$ 4.71 & 6.0257 $\pm$ 0.2451 & 1.6140 $\pm$ 0.1171 \\
            Multi 25 & 0.6036 $\pm$ 0.0055 & 2.57 $\pm$ 0.67 & 9.57 $\pm$ 4.69 & 6.0411 $\pm$ 0.2231 & 1.5649 $\pm$ 0.0930 \\
            Multi 50 & 0.7113 $\pm$ 0.0314 & 1.86 $\pm$ 0.47 & 3.59 $\pm$ 5.93 & 6.0619 $\pm$ 0.2568 & 1.6842 $\pm$ 0.1103 \\
            \bottomrule
        \end{tabular}
    }
    \vspace{+3pt}
    \caption{
    \textbf{Results for Different Watermarks:
    } We report the AUC of our method when watermarking 50\% of the data using Tone- and AudioSeal-based watermarks. We also provide watermark imperceptibility metrics and metrics that capture their impact on model utility.
    }
    \label{tab:table_50_percent}
\end{table}

\subsection{Watermark Detection in Outputs}\label{watermark_detection_in_outputs}
\begin{figure}
    \centering
    \includegraphics[width=0.9\linewidth]{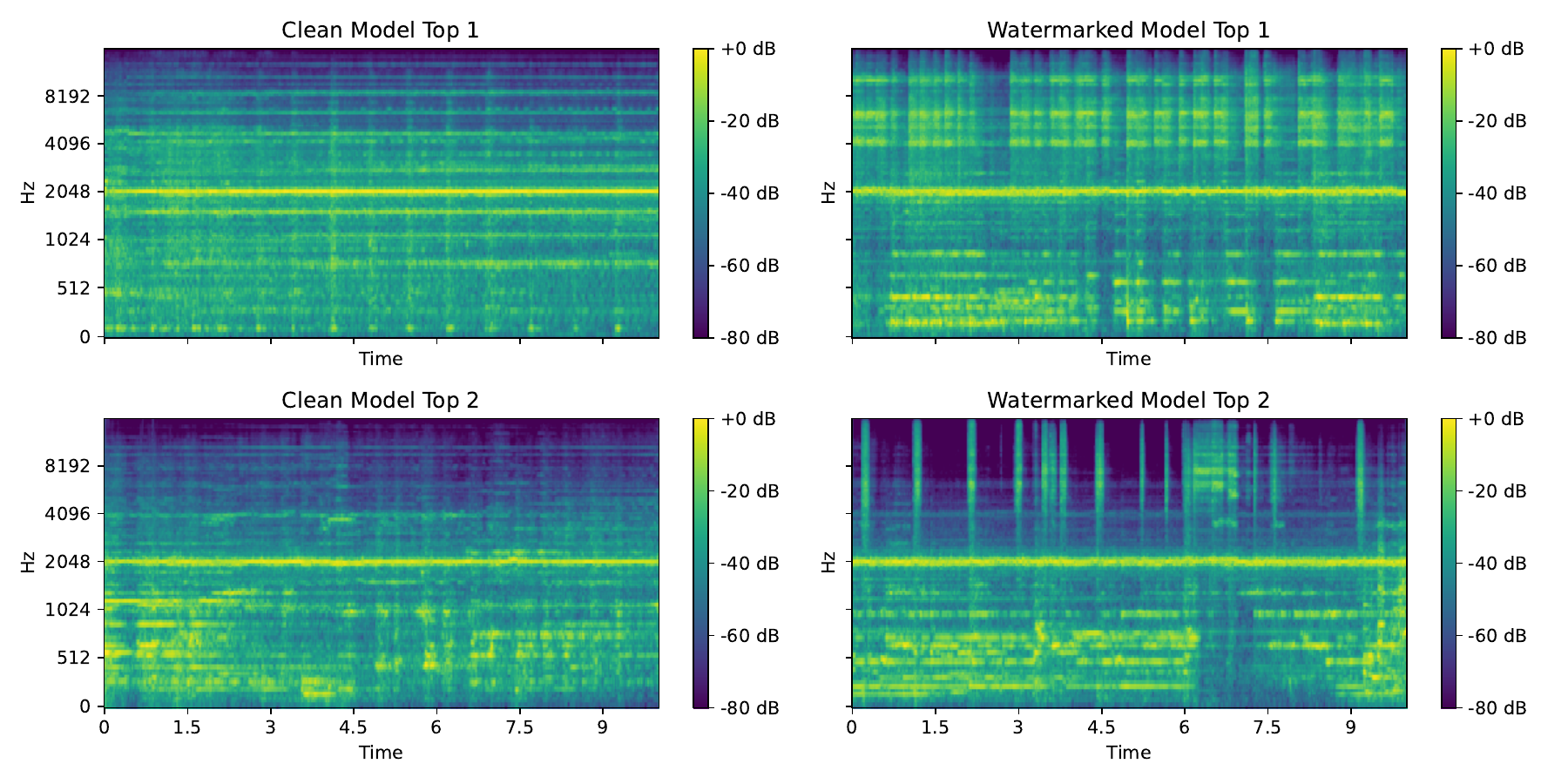}
    \caption{\textbf{Top Audio Continuations} : The mel-spectrogram of the continuations with highest detection scores generated by the clean (left) and watermarked (right) models.}
    \label{fig:top2cont}
\end{figure}

Table \ref{tab:table_50_percent} summarizes the results when adding the considered watermarks into 50\% of the training dataset. First, we focus on the tone-based watermarks. As shown in the table, all of the watermarks significantly outperform random guessing, and we compute the highest AUC (0.7787) with \textit{Tone 440}. However, this result comes with a significant cost of imperceptibility. To this end, we experiment with a watermark in the imperceptible range of frequencies: \textit{Tone 10}. While we compute a drop with respect to the tone in the hearable range, we compute an AUC of 0.7112, indicating the the rule-based classifier still captures a significant different in the generated outputs. Intrigued by the low separability of the generated outputs, we analyze the audio continuations generated by the clean and the watermarked models. Figure \ref{fig:top2cont} plots the two continuations with highest detection according to the rule-based classifier for both models, when injecting \textit{Tone 2048}. The mel-spectrogram reveal that the clean model also generates continuations of the tone. Our results indicate that the watermarked model generates the watermark more often, and with higher intensity.

\begin{figure}
    \centering
    \includegraphics[width=0.7\linewidth]{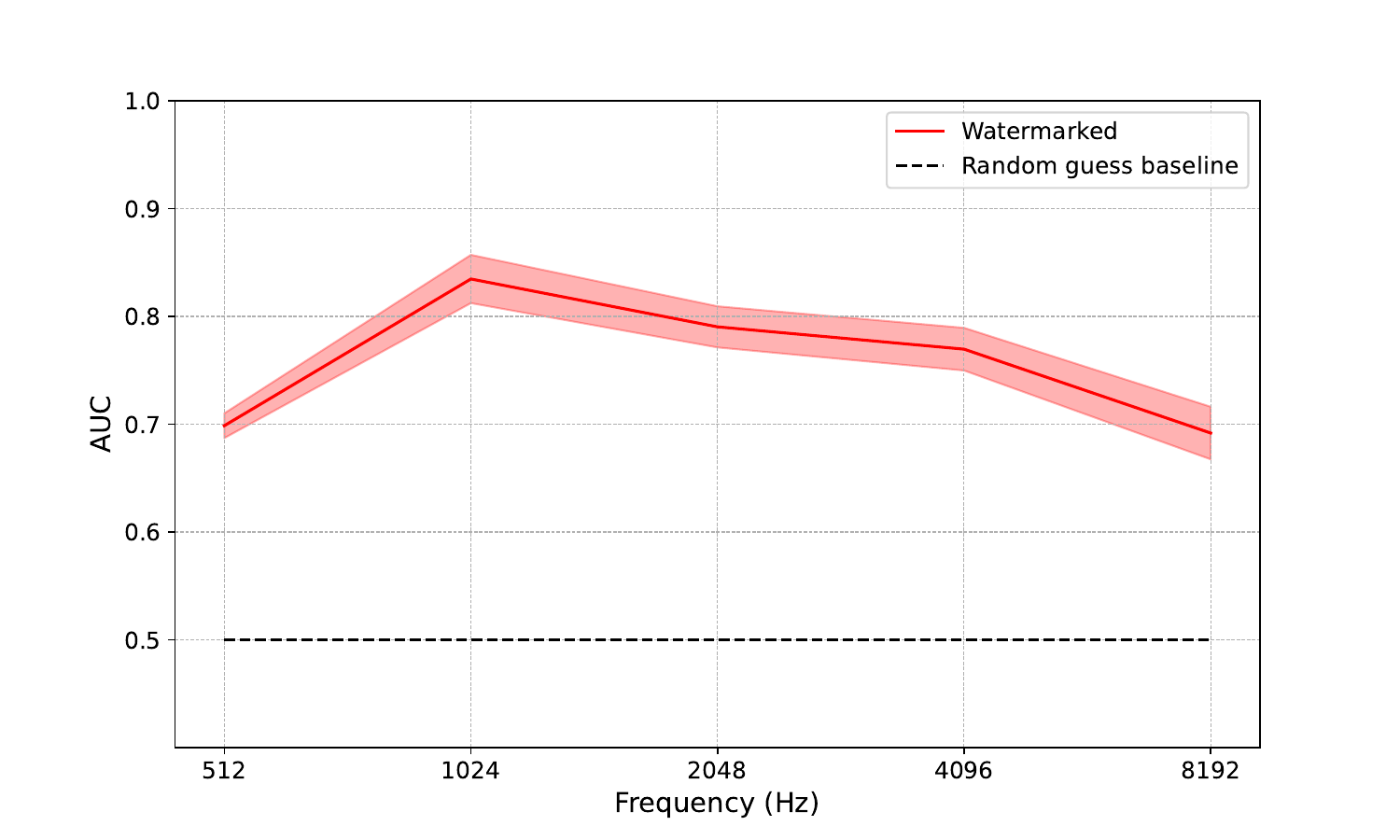}
    \caption{\textbf{AUC for Tones at Different Frequencies: } We plot the AUC of the Rule-based classifier for a wide range of frequencies.}
    \label{fig:tones_512_1024}
\end{figure}

We further investigate the impact of the frequency of the injected tone. Figure \ref{fig:tones_512_1024} plots the results when watermarking 50\% of the data with \textit{Tone f}, for $f = 2^i \text{ Hz}, i \in \{9, 10, 11, 12, 13\}$. For every tone, the rule-based classifier the classifier captures a change in distribution between the outputs produced by the clean model and the watermarked model. We observe the best results when injecting a tone in the band of frequencies between 1024 Hz and 2096 Hz. This band contains frequencies that are more often present in music samples, which can explain why the watermarked model is better at generating these tones.

Next, we experiment with \textit{Secrets}, more specifically \textit{Stop 5} and \textit{Switch 5}, with $f = 440$ Hz and $f' = 880$ Hz. For both watermarks, we obverse a decrease in AUC with respect to the previous watermarks made of continuous tones. The results in Table \ref{tab:table_50_percent} for \textit{Secret 5} indicate that the watermarked model is able to generate the secret frequency. Similarly, the experiment with \textit{Stop 5} suggests that the watermarked model can learn to not generate the continuation of a tone. Moreover, it shows that the watermark does not need to cover the entire audio to still cause a noticeable shift in the generated outputs. 

We now focus on watermarks based on AudioSeal. We experiment with the original AudioSeal watermarking technique, as well as with multiple watermarking levels for $k = 10, 25, 50$. Table \ref{tab:table_50_percent} shows that AudioSeal's detector is not able to outperform random guessing when 50\% of the training data is watermarked with vanilla AudioSeal. We observe an increase in AUC as we increase the number of times we watermark the training samples with AudioSeal. More specifically, we compute the highest AUC of 0.7113 the audios are watermarked 50 times. Our results also show the impact of the watermark's robustness to the model's audio tokenizer. However, all of the AudioSeal-based watermarks that significantly outperform random guessing ($k = 25, 50$) are no longer imperceptible and heavily distort the audios.

Lastly, we assess the impact on model performance of the watermarked data. We compare the model utility of the watermarked models to a clean model by computing the FAD and KLD on the held-out test set. Table \ref{tab:table_50_percent} indicates that every watermarked model performs on par with the clean model across both objective metrics. This indicates that the models are not affected by the degraded training data.

\subsection{Reducing the Amount of Watermarked Data}
\begin{figure}
    \centering
    \includegraphics[width=0.5\linewidth]{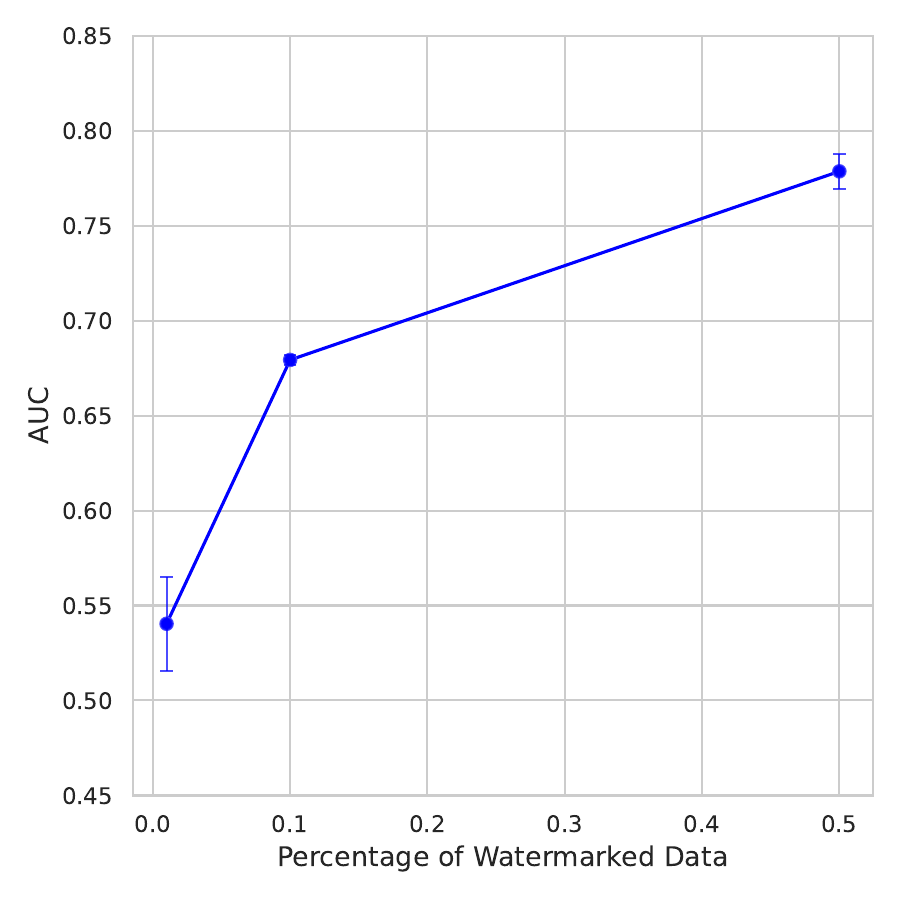}
    \caption{\textbf{Varying Percentage of Watermarked Data : } We plot the AUC when watermarking 1\%, 10\% and 50\% of the data with \textit{Tone 440}.}
    \label{fig:auc_wrt_percentage}
\end{figure}

In a separate experiment, we investigate the impact of the percentage watermarked training data with \textit{Tone 440}. In Figure \ref{fig:auc_wrt_percentage}, we report the AUC of the rule-based classifier when watermarking 1\%, 10\% and 50\% of the data. In all scenarios, the rule-based classifier significantly outperforms the random-guess baseline. The plot shows that the AUC increases for increasing percentages of watermarked data. When watermarking only 10\% of the training data, we compute an AUC of 0.6794.

\subsection{Prompting With Different Tone}

In continuation of our results in Table \ref{tab:table_50_percent}, we investigate whether a model trained on data watermarked with \textit{Tone 440} has become better at continuing any other tone. Using the watermarked model, we generate continuations of audio prompts watermarked with \textit{Tone 880}. We compute an AUC of 0.51, indicating that the rule-based does not outperform random guessing. This finding suggests that the shift in behavior we previously detected is ultimately linked to the specific frequency of the injected tone. 

\subsection{Generating More than One Continuation}
\begin{figure}
    \centering
    \includegraphics[width=\linewidth]{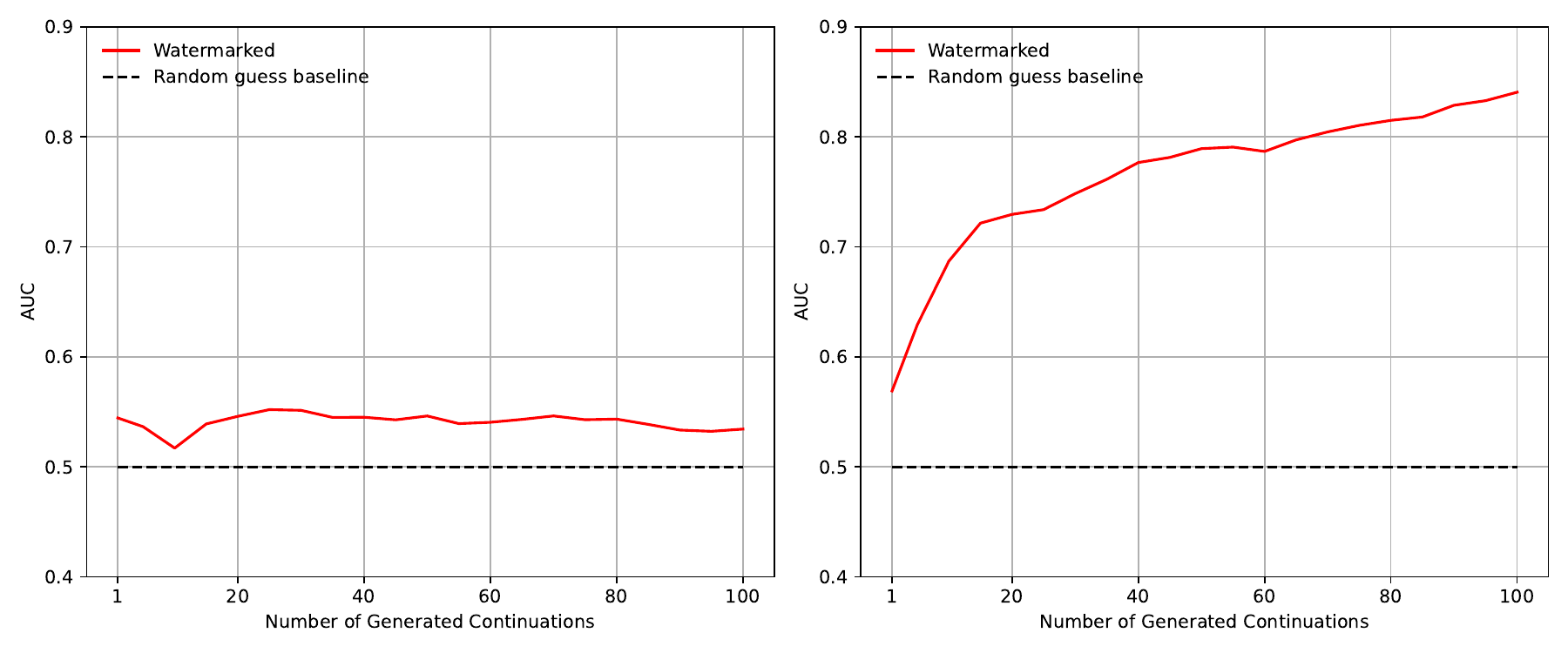}
    \caption{\textbf{Watermark Extraction of Secret:} We plot the AUC of the rule-based classifier for increasing numbers of generated continuations for \textit{Switch 5} in the imperceptible range (left) with $f$ = 5 Hz, $f'$ = 15 Hz, and in the perceptible range (right) with $f$ = 440 Hz and $f'$ = 880 Hz.}
    \label{fig:multiple_continuations}
\end{figure}

We continue experimenting with \textit{Secrets}. Our previous findings suggest that for some \textit{but not all} audio prompts, the watermarked model generates the secret frequency. Motivated by these promising results, we decide to further investigate Switch 5 by generating more than one continuation per model for each sample. We generate up to 100 continuations per sample, and select the continuation with highest prediction score from the rule-based classifier. Figure \ref{fig:multiple_continuations} illustrates the increase in AUC as we increase the number of generated continuations. The improvement in AUC is particularly high when going from 1 to 20 continuations. We compute an increase from 0.569 to 0.7294. Then, the AUC increases at a lower rate, yet without seeming to reach a plateau. This indicates that generating more than 100 continuations could still improve the performance of the rule-based classifier, suggesting that the secret could be extracted from any audio prompt.

We repeat the experiment with \textit{Switch 5} in the imperceptible range of frequency, with $f$ = 5 Hz and $f'$ = 15 Hz. \noindent Figure \ref{fig:multiple_continuations} plots the AUC of the rule-based classifier with respect to the number of generated continuations. The AUC starts at 0.5444 and does not increase as more continuations are generated. This indicates that the secret is not extractable in this setting. We hypothesize that this might be due to the difference between $f$ and the secret frequency $f'$ not being large enough.

\section{Limitations and Future Work}\label{ch:Discussion}

\subsection{Limitations}\label{sec:limit}

While our work provided insights into watermark memorization of a music model, the results are heavily influenced by the training regime of the fine-tuned model. Factors such as the number of epochs and the initial learning rate are known to significantly affect the model's memorization abilities. There is also a potential risk that the data last by the model is more prone to be memorized. To mitigate this, an ablation study could have been conducted to measure the impact of each hyper-parameter on our results, thereby providing a more nuanced perspective. We were limited by time to make a proper ablation study.

\noindent The ideal way to get completely rid of the potential confounding factors of fine-tuning would have been to study the memorization abilities in a clean set up. By having full control over the data ingested and the model architecture (e.g., modifying the audio tokenizer), our results could have been much more meaningful. For example and as previously discussed, the percentage of watermarked data is likely a lower bound to the real percentage it represents. Also, with a clean set up, the watermarked data is completely diluted into the full dataset, removing any bias from its appearance only in the late stages of training. We here discuss the principal bottlenecks for a clean set up. First, our method analyzes the generated content of the models, thus heavily relying on the utility of the model. On one hand, this implies having access to high-quality datasets of textually described music, which are surprisingly scarce. The datasets used to train text-to-music generation models are undisclosed and not freely available. On the other hand, finding the optimal training regime to achieve minimal model utility for our method to work involves a tedious hyper-parameter tuning phase. Moreover, as our analysis covers a plethora of models fine-tuned on different datasets, both compute and storage would have rapidly become intractable.

\noindent Next, our work focuses on one single model, MusicGen. In the initial stages of this work, we identified two possible categories of music generation models to study: models based on diffusion and models with a transformer-based architecture. We discarded diffusion based models to focus on transformer based models, hoping to leverage and transfer the knowledge on LLMs already present in the scientific literature. In the initial phases of this work, we quickly understood that our method would require open access to the training code of the models. We identified two suitable model candidates: VampNet \cite{garcia2023vampnetmusicgenerationmasked} and MusicGen. The organized community around MusicGen, including a dedicated Discord server and available resources, made the decision to select MusicGen straightforward. The fact that MusicGen comes in different model sizes was also convenient. Initially, we thought of studying its impact on memorization, as is often done when studying memorization in LLMs.

\noindent Another limitation of our work is the number of audio watermarks studied. A significant amount of time was spent on identifying suitable watermarking techniques to inject into the training data of the music generation model. In the initial stage of this project, we identified a number of watermarking techniques that might be suited for our study. We had two criteria in mind for the perfect watermarking technique: robustness to the model's audio tokenizer and simple to implement. With Encodec 32 being a very recent compression algorithm, we narrowed down our research on recently developed audio watermarks. Indeed, traditional watermarking techniques, such as those based on Lest Significant Bit, appeared before the release of Encodec 32. Our research narrowed down to two AI-based watermarking techniques that were publicly available and simple to implement: AudioSeal and WavMark \cite{chen2024wavmarkwatermarkingaudiogeneration}. In their work, \cite{audioseal} evaluate the robustness of WavMark to Encodec 32 and show that it is not robust to it. 

\subsection{Future Work}

There is a potential in research on the topic of memorization in audio generation models. Current research is more focused towards text and image generation model, leaving a gap yet to be explored in the audio domain. We here propose future angles of research in the continuation of our work.

\noindent As discussed in \ref{sec:limit}, it would be interesting to evaluate our method on a different model than MusicGen. Especially, our results indicate that the audio tokenizer has significant impact on our method. In this spirit, we hope that MusicGen with the audio tokenizer based on DAC will be released.

\noindent Future experiments also include the evaluation of our method with different watermarking techniques. This about, the training code of AudioSeal has been recently released. This enables to train a watermarking technique that is more robust to Encodec 32 than the current released version.

\section{Conclusion}\label{ch:conclusion}

The goal of this work was to develop a method for detecting copyrighted content in the training data of music generation models. To achieve this, we experimented with watermarking the training data of MusicGen, a state-of-the-art music generation model, using different audio watermarking techniques.

\noindent We have identified two factors that influence the success of our watermark injection method. 

\noindent The first factor is the amount of watermarked data that gets injected into the training data of MusicGen. Our method performs better when more watermarked samples are present in the fine-tuning dataset. This finding aligns with current research on LLMs, where it has been shown that duplicated sequences are more prone to be memorized \cite{kandpal2022deduplicatingtrainingdatamitigates}. Our results indicate that injecting a watermark made out of a perceptible tone into 10\% of the fine-tuning data of MusicGen leads to a detectable shift in the samples generated by the model. 

\noindent The second factor that contributes to the success of our method is the robustness of the watermark to MusicGen's audio tokenizer. Our experiments with AudioSeal and multiple watermarking indicate that higher robustness to the tokenizer results in better model attribution from the outputs. When injecting audios watermarked more than 25 times with AudioSeal, our results suggest that there is significant difference in the outputs generated by the clean and the watermarked model. Moreover, this also proves the model's ability to learn more complex and unique watermarks, such as those generated by AudioSeal. These results also show the need to evaluate our method using different audio tokenizers. We further discuss this in \ref{sec:limit}.

\noindent Our results also show that MusicGen's outputs can be influenced by injecting watermarks in the range of frequencies imperceptible to the human ear. Indeed, the outputs of a model where 50\% of the training data contains a tone at 10 Hz significantly differ from those of a clean model. The imperceptibility of this watermarking technique is supported by high PESQ scores. These tones are however subject to removal through high-pass filtering.

\noindent From our experiments on secrets in the perceptible range of frequencies, we showed that it is possible to improve our method's effectiveness by generating more than one continuation per prompt. Our results suggest that the secret can potentially be extracted from any prompt, provided that we generate enough outputs from it. However, constrained by the range of imperceptible frequencies (0 Hz to 20 Hz) that can be injected into the model's training data, we were not able to reproduce a secret with the same properties in the imperceptible range.

\noindent Moreover, we acknowledge that our results are highly influenced by the model's training regime. Since we did not control for different parameters, such as the learning rate or the number of epochs, our results should not be considered in absolute terms. Nevertheless, we believe that the relationships we discovered still provide meaningful insights into the impact of watermarks on the outputs of music generation models.

\bibliographystyle{plainurl}
\bibliography{References}



\end{document}